\documentclass[letterpaper, 10 pt, conference]{ieeeconf} 
\IEEEoverridecommandlockouts
\usepackage{url}
\usepackage{float}
\usepackage{amsmath}
\usepackage{graphicx}
\usepackage{subcaption}
\usepackage{siunitx}
\usepackage{textcomp}
\usepackage{algorithm}
\usepackage{algorithmicx}
\usepackage{algpseudocode}
\usepackage{bm}

\usepackage{multicol}
\usepackage{lipsum}
\usepackage{mwe}

\usepackage[normalem]{ulem}

\usepackage{fancyhdr}

\newcolumntype{C}[1]{>{\centering\arraybackslash}m{#1}}
\usepackage[thinlines]{easytable}
\usepackage[colorlinks = true,
            linkcolor = blue,
            urlcolor  = blue,
            citecolor = blue,
            anchorcolor = blue]{hyperref}

\begin{document}
\title{End-To-End Real-Time Visual Perception Framework for Construction Automation}

\author{Mohit Vohra$^1$, Ashish Kumar$^1$, Ravi Prakash$^1$ and Laxmidhar~Behera$^{1,2}$,~\IEEEmembership{Senior Member,~IEEE} 
\thanks{$^1$Authors are with Indian Institute of Technology Kanpur, India,\newline
\{\texttt{mvohra, krashish, ravipr and lbehera\}@iitk.ac.in}.\newline
$^2$ Author is with TCS Innovation Labs, Noida, India}}

\maketitle

\begin{abstract}
In this work, we present a robotic solution to automate the task of wall construction. To that end, we present an end-to-end visual perception framework that can quickly detect and localize bricks in a clutter. Further, we present a light computational method of brick pose estimation that incorporates the above information. The proposed detection network predicts a rotated box compared to YOLO and SSD, thereby maximizing the object's region in the predicted box regions. In addition, precision (\textit{P}), recall (\textit{R}), and mean-average-precision (mAP) scores are reported to evaluate the proposed framework. We observed that for our task, the proposed scheme outperforms the upright bounding box detectors. Further, we deploy the proposed visual perception framework on a robotic system endowed with a UR5 robot manipulator and demonstrate that the system can successfully replicate a simplified version of the wall-building task in an autonomous mode.
\end{abstract}

\section{Introduction}
Manufacturing and construction are one of the widespread and continuously growing industries. The former has seen a dramatic increase in production capacity due to the optimization of industrial automation, while the latter has adopted automation only marginally \cite{asadi2018real}. Construction automation is inherently quite challenging for several reasons. First, the workspace is highly unstructured. Therefore, very high precision and robust visual perception, motion planning, and navigation algorithms are required for autonomous solutions to adapt to different scenarios. Secondly, a mobile manipulator needs to move between multiple positions, compelling us to perform onboard computations for various algorithms. Therefore, limited memory, power, and computational resources make this task more challenging.

The process of automation can have a broad impact on the construction industry. First, construction work can continue without pause, which ultimately shortens the construction period and increases economic benefits. Also, the essential benefits of construction automation are worker safety, quality, and job continuity. Towards this end, recently, Construction Robotics, a New York-based company, has developed a bricklaying robot called SAM100 (semi-automated mason) \cite{parkes2019automated}, which makes a wall six times faster than a human. However, their robot required a systematic stack of bricks at regular intervals, making this system semi-autonomous, as the name suggests.

One of the primary construction tasks is to build a wall from a pile of randomly arranged bricks. To replicate the simplified version of the above work, humans must complete a sequence of operations: \textit{i)} Select the appropriate brick from the pile, e.g., the topmost brick, \textit{ii)} finding the optimal grasp pose for the brick, and \textit{iii)} finally, placing the brick in its desired place, i.e., on the wall. Humans can do this work very quickly and efficiently. However, the robot must perform a complex set of underlying operations to complete the above steps autonomously \cite{prakash2019learning}.

\begin{figure}[t]
  \centering
      \begin{subfigure}[b]{0.45\linewidth}
    \includegraphics[width=0.965\linewidth,height=0.750\linewidth]{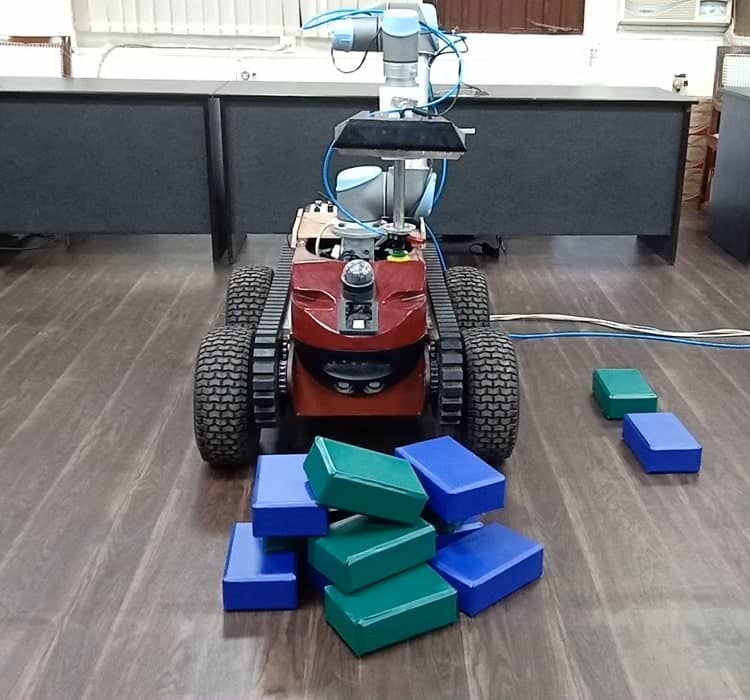}
    \caption{}
    \label{fig:hardware_setup}
  \end{subfigure}
  \hspace{2ex}
  \begin{subfigure}[b]{0.45\linewidth}
    \includegraphics[width=0.965\linewidth,height=0.750\linewidth]{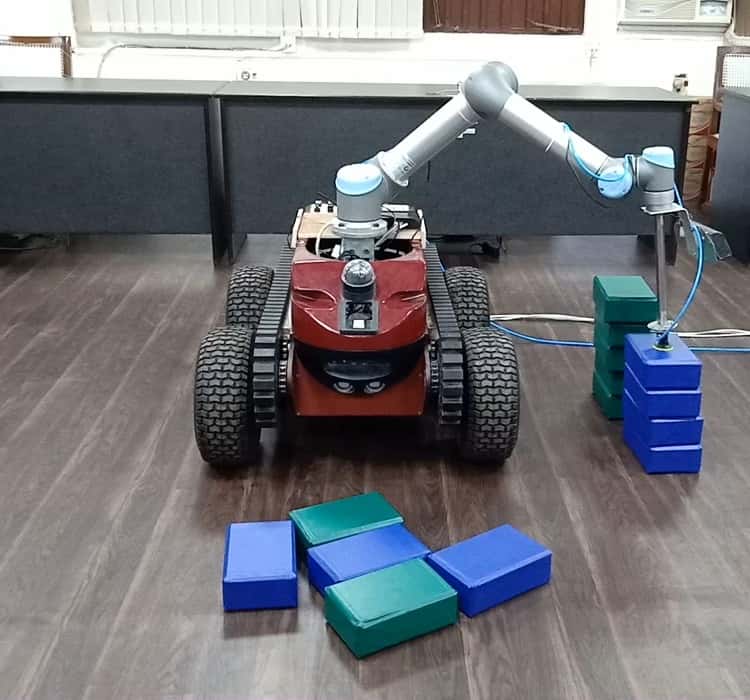}
    \caption{}
  \end{subfigure}
  \caption{ 
          (a) shows a simple scenario where a pile is located near the robotic system, (b) the robotic system mimics wall building task i.e. detects pile, selects a target brick and constructs a wall on its side in a fully autonomous way.
            }

  \label{fig: task}  \vspace{-4mm}

\end{figure}

In this paper, we aim to deploy a robotic solution for the task of construction automation in a constrained environment (Fig.\ref{fig: task}) with limited computational resources (single CPU with I7 processor, 4core, 8GB RAM ). We assume that all bricks are of equal size, and their dimensions are known. We further assume that the wall assembly area and brick piles are very close, exempting us from deploying any localization and navigation modules for robots. Thus, the main challenge in this task is to detect and localize bricks in the clutter while handling the multiple instances of the bricks. Once we have localized the bricks in a clutter, we will use the above information to estimate the brick pose. Following are the main contributions in this paper:
\begin{itemize}
    \item A computationally efficient object detection network for the detection and localization of bricks in a clutter is presented in this paper.
    \item A light computational method for estimating brick pose using point cloud data is presented in this paper.
    \item All the modules are integrated into a robotic system to develop a fully autonomous system.
    \item Extensive experiments to validate the performance of our system.
\end{itemize}{}

In the next section, we briefly provide a review of state-of-the-art algorithms related to the paper. In the section-\ref{sec_prob}, we formulate the problem statement. The overall approach and its modules are explained in section-\ref{sec_network}. In section-\ref{sec_exp}, the experimental study of the algorithm is reported for various test cases. This paper is finally concluded in section-\ref{sec_con}.
\section{Related Works}
\label{sec_background}
\subsection{Object Detection}

As mentioned in the previous section, the first stage of the construction process is the localization of the target object. In our case, the target object is referred to as a brick. In general, bricks are arranged randomly. Therefore, the brick must be localized before grasping. The process of brick localization falls under the category of the object detection algorithm. We perform a brick localization process in the image space. Several object detection algorithms exist in the literature. Here, we limit our discussion to only Conventional Neural Network (CNN) based methods.

The RCNN \cite{RCNN} generates object proposals (rectangular regions) in the image plane, and a CNN is used to extract features from the proposed regions, followed by a classifier to classify the proposed regions into N different classes, where N may vary according to application. In RCNN, most of the time is consumed in proposal generation as this step is performed on CPU, and also inference time increases linearly with an increase in the number of proposals. SPPnets \cite{SPPNet} were proposed to speed up the RCNN by extracting the features for the whole image at once and then cropping the feature map corresponding to the proposals. Due to the multistage nature of the above algorithm, joint training was required. Fast-RCNN \cite{Fast-RCNN} proposes an improved approach that is relatively faster and requires single-stage training. A further improved version of Faster-RCNN \cite{Faster-RCNN} was also proposed in which proposals are generated within the CNN, called a Region Proposal Network (RPN). The RPN was the key improvement in improving the overall algorithmic real-time performance. All the methods discussed above predict an upright bounding box around the detected object. In addition to the target object region, the predicted box may contain non-object regions or backgrounds. Hence,  to minimize the background in the detected boxes, various solutions are present in the literature. For example, in \cite{li2018multiscale}, the author predicts a rotated bounding box from the set of prior rotated boxes (anchor boxes). Similarly, Mask-RCNN \cite{Mask-RCNN} can predict the bounding box and mask of the object simultaneously, which is known as instance detection and segmentation.

All the algorithms mentioned above consists of two steps; \textit{i)} generation of object proposals or anchor boxes (axis-aligned or rotated), \textit{ii)} classification (or regressing) the proposals using a CNN with a backbone such as VGG \cite{vgg}, ResNet \cite{resnet}. Thus the performance of the algorithm depends on the proposal generation process. On the other hand, authors of You Only Look Once (YOLO-v1) \cite{yolov1} have proposed a single network for object detection which divides the image into grid cells and directly predicts the fixed number of bounding boxes, corresponding confidence scores, and class probabilities for each grid cell. In the same direction, single shot multibox detector (SSD) \cite{SSD} is another variant of a single-stage object detector. In this variant, multi-resolution object detection is performed, i.e., detecting the presence of an object and its class score at various stages of different spatial resolutions.

\subsection{6D Pose Estimation}
After brick localization, a grasp operation needs to be performed by the manipulator. Choosing an optimal grasp configuration is a non-trivial task and remains an open problem. The grasp configuration depends on the 6D pose of the brick. Several Neural-network based pose estimation methods exist in the literature \cite{tekin2018real}, \cite{he2020pvn3d}, but limited memory resources compel us to use computationally light pose-estimation methods. To this end, several algorithms exist to estimate the object's 6D poses, which require a high degree of surface texture on the object. In our case, the estimation of brick poses is quite challenging due to their cubic shape (flat surfaces), which do not have surface textures. Therefore, the feature point matching technique \cite{vfh} \cite{do2016efficient} cannot be used. Other approaches \cite{drost2010model}, \cite{hinterstoisser2016going} consider an earlier model of the object. These methods require a preprocessing step, followed by the correspondence matching process. The matching process is the most time-consuming component of such algorithms. Besides, point-to-point matching methods (ICP \cite{icp}, GICP \cite{gicp}) are based on local geometric properties. Therefore, these methods can be stuck in local minima when aligning the target model with the reference due to flat surfaces.
\section{Problem Statement}
\label{sec_prob}

\subsection{Object Detection}
\begin{figure}[!h]
  \centering
      \begin{subfigure}[b]{0.33\linewidth}
    \includegraphics[width=0.965\linewidth]{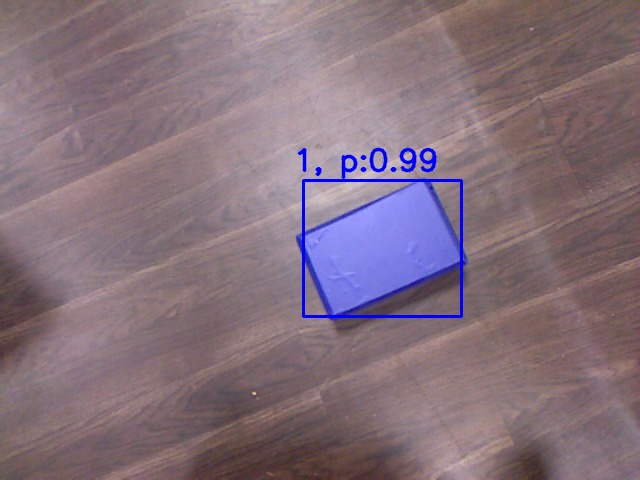}
    \caption{Predicted Boxes}
    \label{fig:nms_prob_a}
  \end{subfigure}%
  \begin{subfigure}[b]{0.33\linewidth}
    \includegraphics[width=0.965\linewidth]{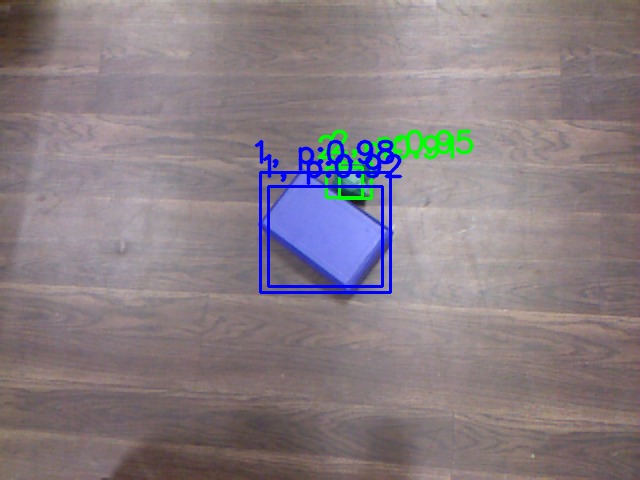}
    \caption{Boxes Overlap}
    \label{fig:nms_prob_b}
  \end{subfigure}%
  \begin{subfigure}[b]{0.33\linewidth}
    \includegraphics[width=0.965\linewidth]{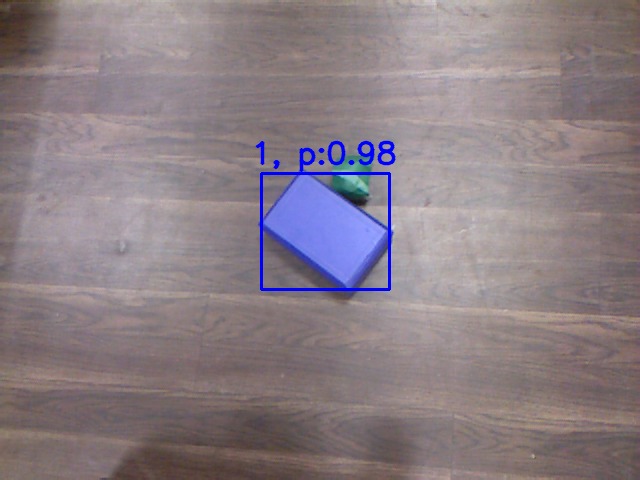}
     \caption{After NMS}
     \label{fig:nms_prob_c}
  \end{subfigure}
  \caption{Nearby predictions can be missed due to NMS}
  \label{fig:nms_prob}
\end{figure}

As mentioned in previous sections, the main challenge is to identify the bricks in a clutter. All state-of-the-art object detectors predict the upright or straight bounding box, which has three limitations:
\begin{itemize}
    \item The bounding box corresponding to the rotated or tilted object contains a  significant non-object region (Fig. \ref{fig:nms_prob_a}). Thus, it requires an additional step to extract object information, like object segmentation in the bounding box region.
  
    \item If two or more objects are very close to each other. The corresponding bounding boxes will have non-zero intersecting regions (Fig. \ref{fig:nms_prob_b}). Thus, additional steps are required to handle intersecting regions, as this region may contain clutter for one box or an object part for another box.
    
    \item If the intersecting regions are significant, after applying non-maximal-suppression (NMS), neighbor detection may be missed \cite{nms_prob}, as shown in Fig. \ref{fig:nms_prob_c}.
    
\end{itemize}{}

To compete with the above limitations, we designed a CNN-based model to detect and localize bricks by predicting rotated boxes. An additional degree of freedom, i.e., the box's angle, allows the network to predict the box with greater alignment with the target object. Since most of the area inside the rotated bounding box corresponds to the target object, we can directly use the region corresponding to the rotated bounding box to extract the target object's information, avoiding any additional computations. A detailed description of the proposed model is given in Section-\ref{sec_network}.

\subsection{6D Pose Estimation}
As mentioned earlier, the bricks used in the experiments have a flat and textureless surface. Therefore, feature matching methods for pose estimation are unreliable in our experiments, as the number of features is less and not very distinct. Since the bricks used in our experiments have a cuboidal shape, if we can estimate the pose of at least one surface of the brick, this information is sufficient to estimate the pose of the entire brick. The brick has six faces, and each face has a specific relative pose with a local brick frame. Hence to estimate the brick pose, we have to identify the brick surface (out of six surfaces), estimate the surface pose, and use relative transformations to get the complete brick pose. A brief description of the pose estimation method is given in the section - \ref {sec_pose}.
\section{The Proposed Framework}
\label{sec_network}

\begin{figure}[t]
\centering
  \begin{subfigure}[b]{0.45\linewidth}
    \includegraphics[width=\linewidth,height=0.65\linewidth]{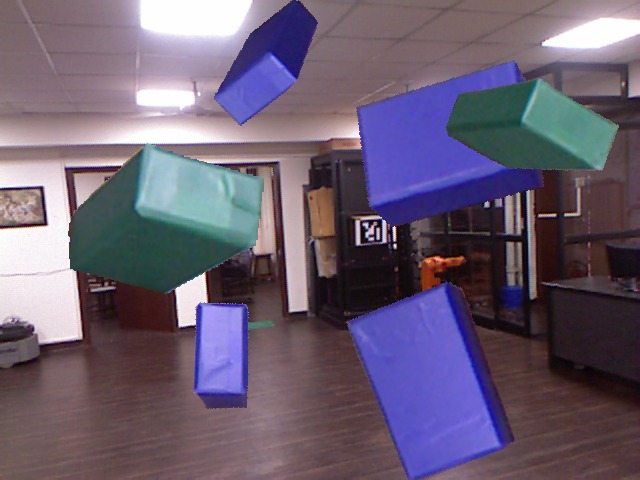}
  \end{subfigure}
    \begin{subfigure}[b]{0.45\linewidth}
    \includegraphics[width=\linewidth,height=0.65\linewidth]{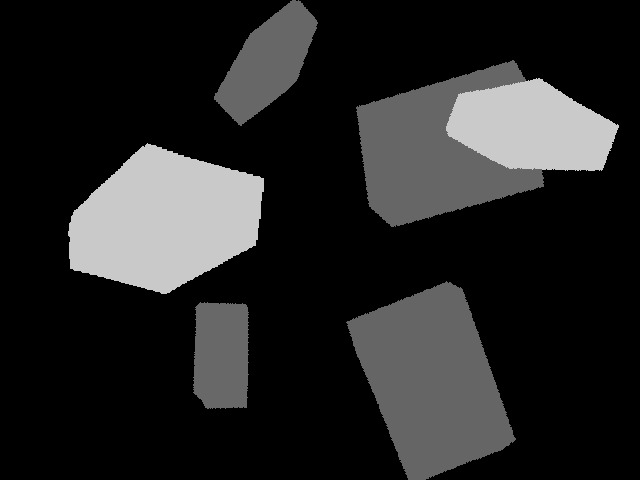}
  \end{subfigure}
  \caption{A synthetically generated sample}
  \label{fig:random sample data}
  \vspace{-5mm}
\end{figure}

\subsection{Dataset}
We collect $30$ images for each class of brick (e.g., blue and green) using Kinect. Images are collected, such that each image has an isolated brick with a different pose. A manual mask is generated for each image. Following \cite{kumar2019semi}, the raw images and the corresponding masks are used to synthesize the cluttered scenes. A dataset of $10$k training images and $5$k test images are generated. We generate ground truths for synthetic images such that a unique instance-ID, as opposed to semantic segmentation, is assigned for each brick instance. A random sample from the data set is shown in Fig. \ref{fig:random sample data}. Furthermore, for each mask instance, we generate a rotated box using the OpenCV API.

Each image is divided into several grids, where each grid has a size of $16\times16$ pixels. Thus if the raw image has a size of $480\times640$, then the total number of the grids are $\frac{480}{16} \times \frac{640}{16}$, i.e., $30\times40$. Further, each grid is represented by an $8D$ vector representing the three-class probabilities (blue brick, green brick, or background) and five bounding box parameters $x, y, w, h, \theta$. For each grid, class probabilities are assigned if the rotated bounding box's centroid (corresponding to the blue brick, green brick) exists within that grid. If there is no centroid in the grid, then we will assign the probability of $1.0$ to the background label. Suppose a centroid exists within a grid. Corresponding bounding box parameters are $x, y, w, h, \theta$, where $x, y$ is the offset between the rotated bounding box center and the topmost grid corner. Parameters $w, h, \theta$ are width, height, and orientation of the bounding box, respectively. We scale the bounding box parameters in the range of ($0, 1$), where the maximum value of offset is $16$ pixels. The box's maximum dimension can be $480 \times 640$ pixels, and the maximum orientation value is $3.14$ radians. Thus for each image, we have an output tensor of size $30\times40\times8$. Further, if multiple centroid points exist in a single grid, we select the centroid point corresponding to the mask, which has a larger fraction of the complete mask in that grid.

\begin{figure}
  \includegraphics[width=\linewidth]{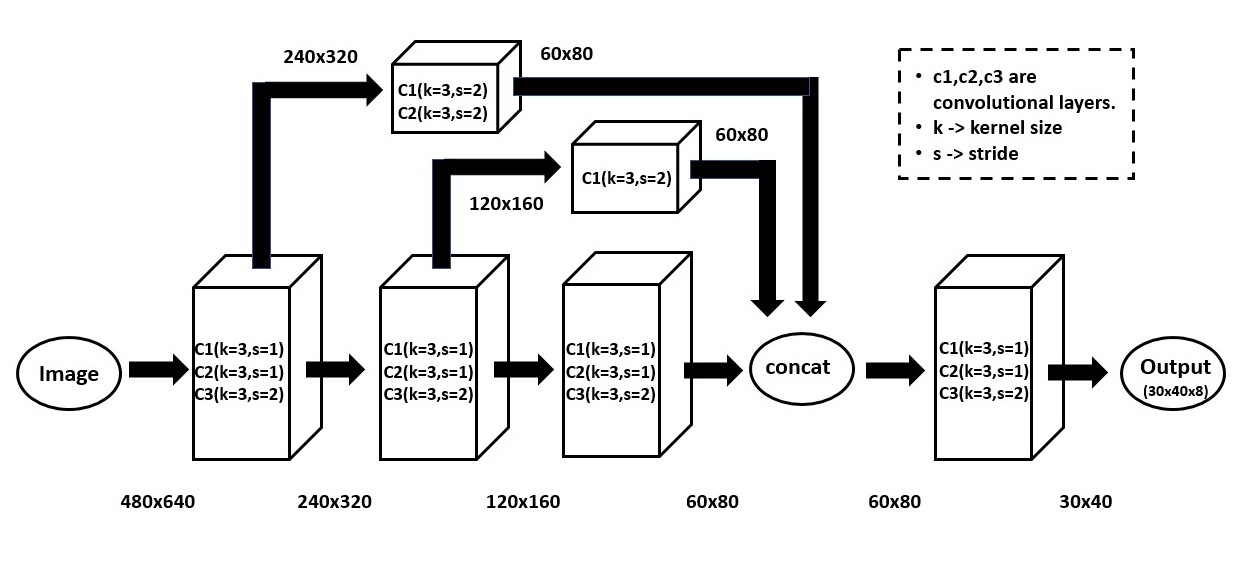}
  \caption{Proposed Network}
  \label{fig:network}
  \vspace{-5mm}
\end{figure}

\subsection{Rotating Box Network}
Fig. \ref{fig:network} represents the network architecture. For each block, the size of the input and output feature map is mentioned. Also, we use the ReLU activation function after each layer. The SSD architecture inspires the proposed rotating box network architecture. In SSD, shallow layer features and depth layer features are used for final predictions. Similarly, in the proposed network, features from individual shallow layers are processed, concatenated, and passed through a series of fully convolutional layers for final prediction. Unlike SSDs, the proposed network prediction does not use any anchor boxes. Instead, it predicts an additional degree of freedom (angle of the box), and thus the predicted bounding boxes can align more accurately than the constrained bounding box.

\begin{figure}[h]
\begin{subfigure}[b]{0.11\textwidth}
                \centering
             \includegraphics[scale=0.08]{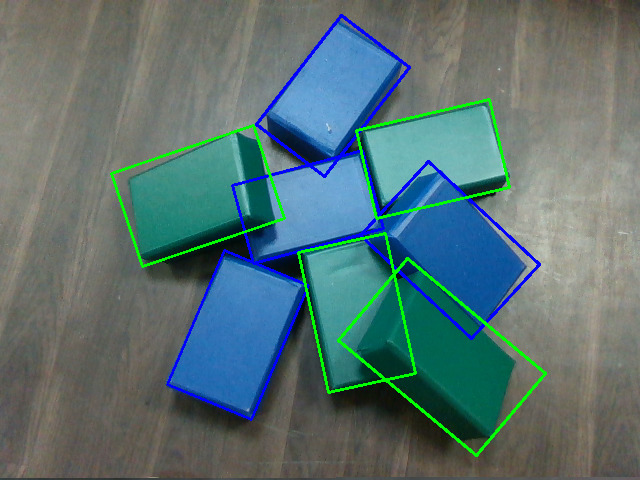}
        \end{subfigure}%
        \begin{subfigure}[b]{0.11\textwidth}
                \centering
                \includegraphics[scale=0.08]{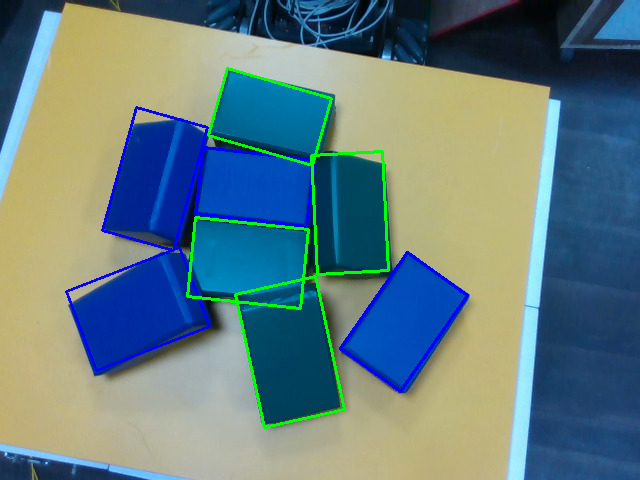}
        \end{subfigure}%
        \begin{subfigure}[b]{0.11\textwidth}
                \centering
                \includegraphics[scale=0.08]{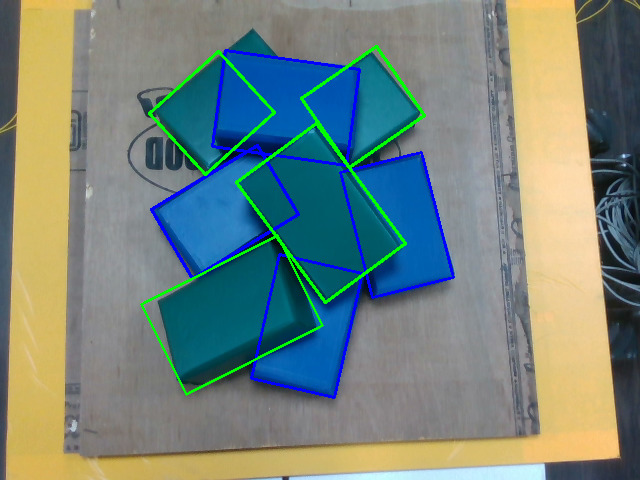}
        \end{subfigure}%
        \begin{subfigure}[b]{0.11\textwidth}
                \centering
                \includegraphics[scale=0.08]{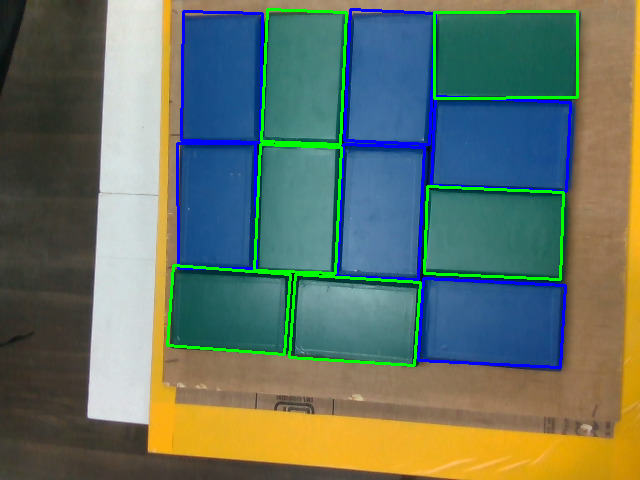}
        \end{subfigure}
        \caption{Network Predictions}
  \label{fig:all_all_results}
  \vspace{-4mm}
\end{figure}

In order to train the network for predicting rotated boxes, the input to the network is the raw image, and the output of the network is a tensor of size $30 \times 40 \times 8$. Further, we use a cross-entropy loss for the class probabilities and a regression loss for the bounding box parameters. Overall loss for the network is the average of both losses. Further, to avoid any biasing in training because of the large number of non-object grids as compared to the object grids, we select positive to negative ratio = $1:2$ by following \cite{SSD}. Output the model for the different arrangement of bricks in a variety of backgrounds is shown in Fig. \ref{fig:all_all_results}.

\subsection{Pose Estimation}
\label{sec_pose}
\begin{figure}[h]
    \centering
\begin{subfigure}{0.125\textwidth}
  \includegraphics[width=1.18\linewidth]{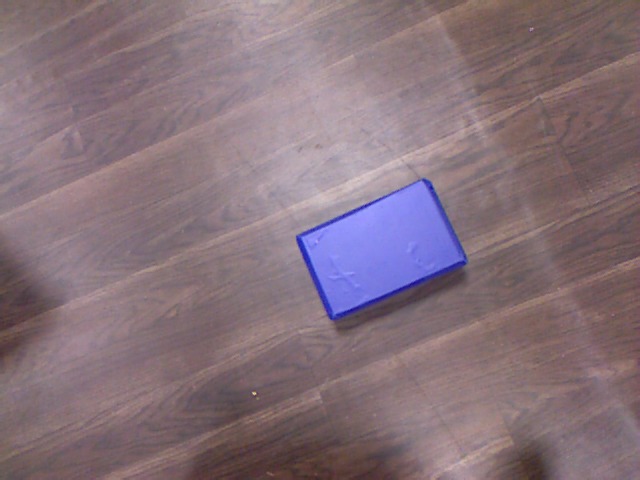}
  \caption{Image}
  \label{fig:pose1}
\end{subfigure}\hfil
\begin{subfigure}{0.125\textwidth}
  \includegraphics[width=1.18\linewidth]{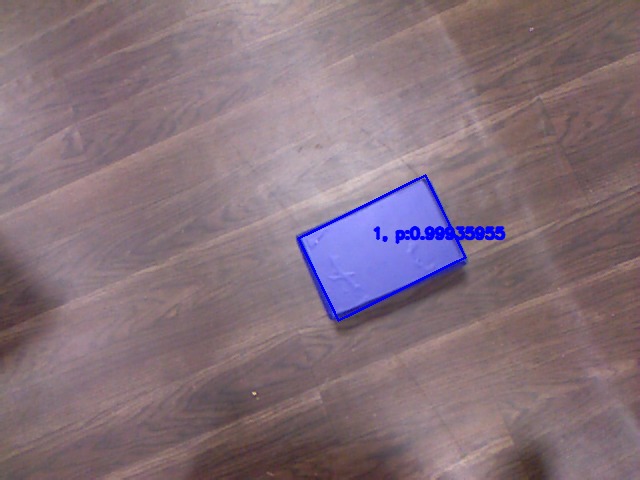}
  \caption{Rotating box}
  \label{fig:brick_seg}
\end{subfigure}\hfil
\begin{subfigure}{0.125\textwidth}
  \includegraphics[height=0.89\linewidth, width=1.15\linewidth]{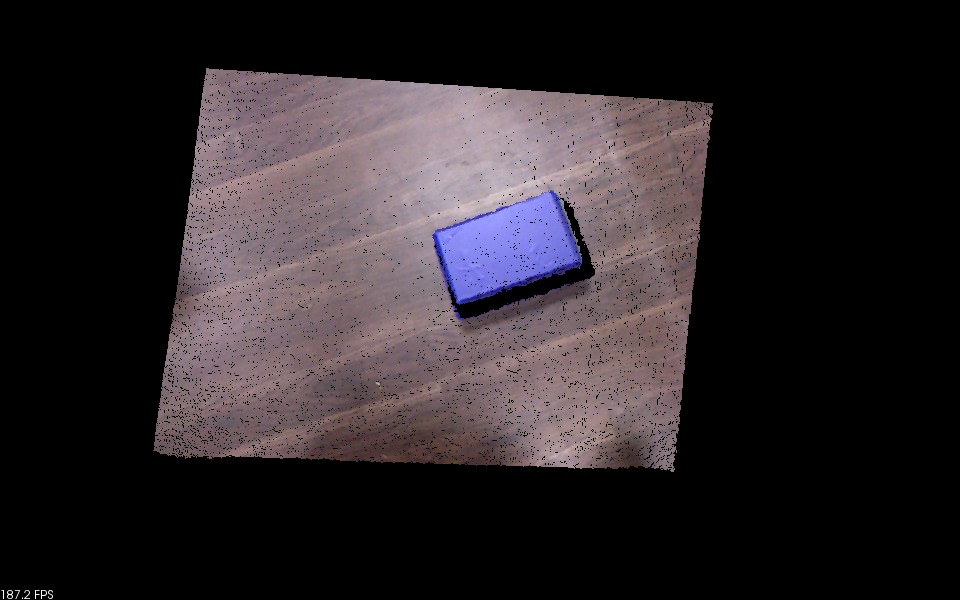}
  \caption{Point Cloud}
\end{subfigure}

\medskip
\begin{subfigure}{0.125\textwidth}
  \includegraphics[height=0.89\linewidth, width=1.16\linewidth]{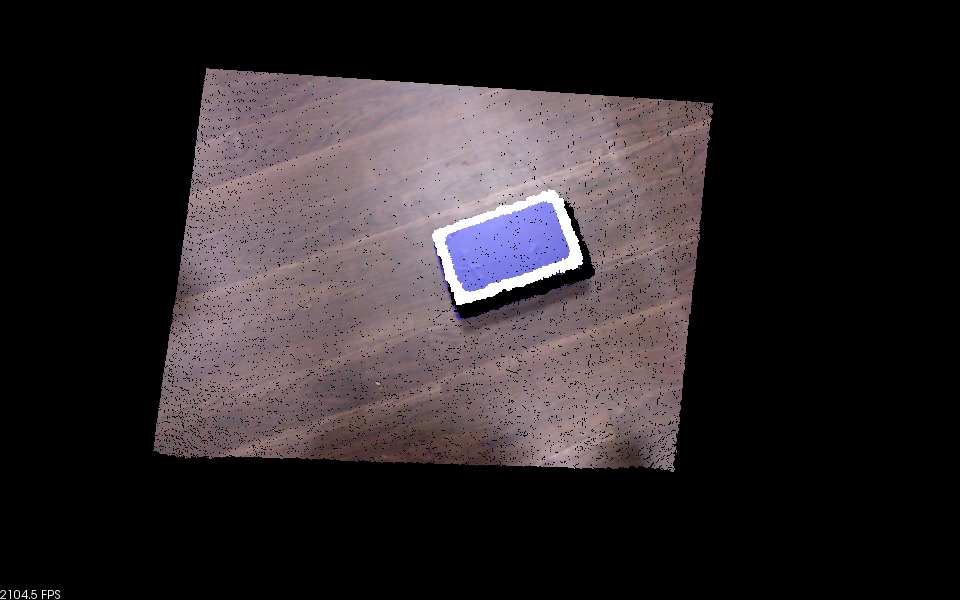}
  \caption{Edges}
  \label{fig:edges}
\end{subfigure}\hfil
\begin{subfigure}{0.125\textwidth}
  \includegraphics[height=0.89\linewidth, width=1.16\linewidth]{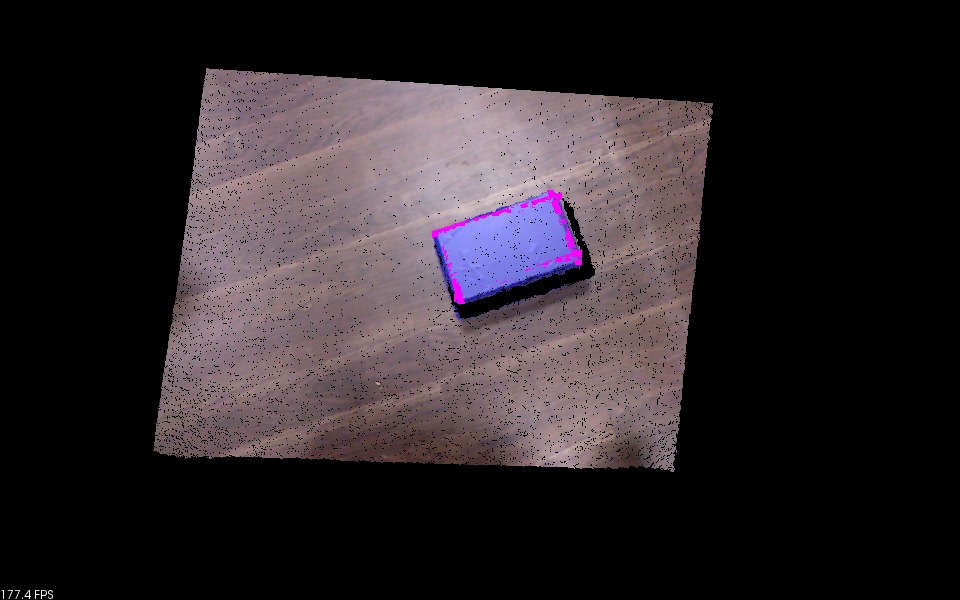}
  \caption{Lines}
  \label{fig:lines}
\end{subfigure}\hfil
\begin{subfigure}{0.125\textwidth}
  \includegraphics[height=0.89\linewidth, width=1.15\linewidth]{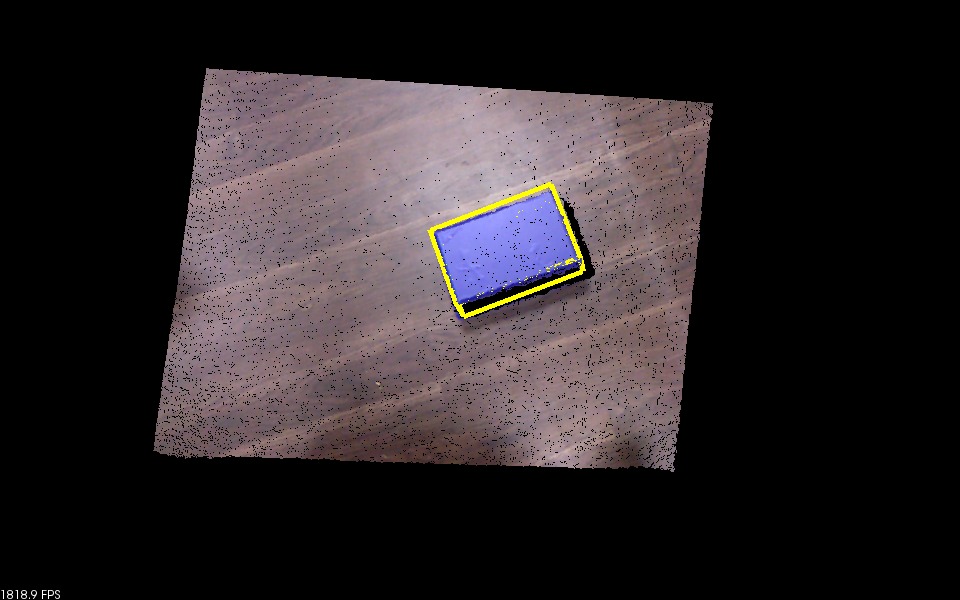}
  \caption{6D Pose}
  \label{fig:pose}
\end{subfigure}
\caption{Pose estimation pipeline}
  \label{fig:complete_pose}
    \vspace{-4mm}
\end{figure}

To estimate the brick pose, as mentioned earlier, we have to calculate the pose of one of the brick surfaces and use relative transformation to extract the complete brick pose. For this task, we feed the current image (Fig. \ref{fig:pose1}) to the rotating box network. The region corresponding to the rotating box (Fig. \ref{fig:brick_seg}) is called the brick region, and the point cloud corresponding to the brick region is called the brick cloud. On the brick's cloud, we apply the following steps:

\begin{itemize}
    \item Apply \textit{RANSAC} method for estimating a set of points (inliers)  that fits a planar surface in the brick cloud data.
    \item Compute the centroid, major axis, and minor axis of the inliers. Together these three pieces of information represent the pose of the planar surface. To estimate the surface ID, we follow the following steps.
    \item Using \cite{vohra2019real}, extract all boundary points in the inliers, which is marked in white color in Fig. \ref{fig:edges}.
    \item Apply RANSAC method for fitting the lines on the boundary points which are shown pink in Fig. \ref{fig:lines}.
    \item Compute all corner points, which are the intersecting point of two or more lines.
    \item Pair the corner points representing the same line \cite{icinco21}, and the distance between two corner points gives the length of the edge.
    \item Since the brick dimensions are known in advance. Hence the length of the edges can be used to identify the surface, and we can use relative transformation to compute the 6D pose of the brick, as shown in Fig. \ref{fig:pose}.
\end{itemize}

\section{Experiments and Result}
\label{sec_exp}

\subsection{Experimental Setup}
For experimental evaluation, we use our robotic platform setup, as shown in Fig. \ref{fig:hardware_setup}. It consists of a UR5 robot manipulator with its controller box (internal computer) mounted on a ROBOTNIK Guardian mobile base and a host PC (external computer).  The UR5 robot manipulator is a 6-DOF robotic arm designed to work safely alongside humans. We use an eye-in-hand approach, i.e., the image acquisition hardware, which consists of RGB-D Microsoft Kinect Sensor, is mounted on the manipulator. A suction-based gripper is used for grasping. Robot Operating System (ROS) is used to establish a communication link among the sensor, manipulator, and the gripper.

\begin{figure*}
 \begin{subfigure}[b]{0.14\textwidth}
                \centering
                \includegraphics[scale=0.08]{images/experiments/1.jpg}
                \caption{}
                \label{fig:full_system1}
        \end{subfigure}%
        \begin{subfigure}[b]{0.14\textwidth}
                \centering
                \includegraphics[scale=0.08]{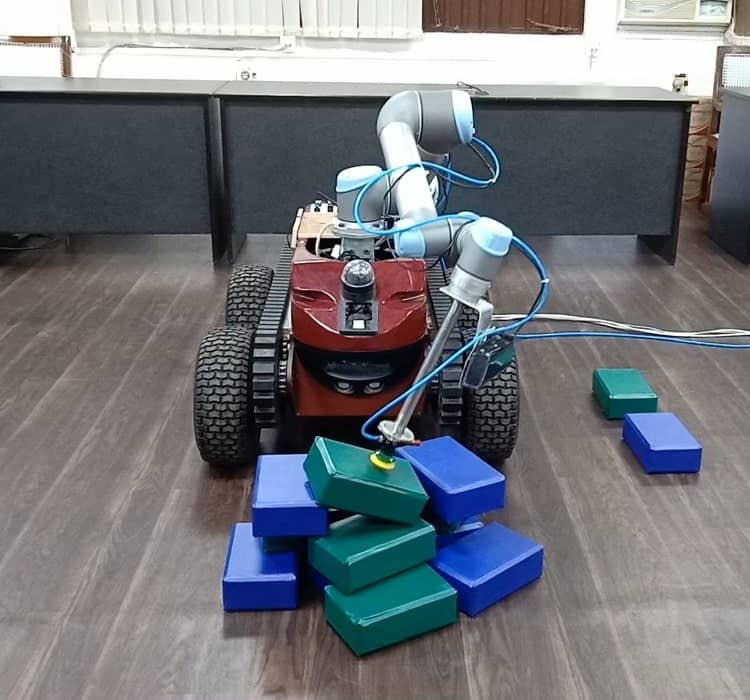}
                \caption{}
                \label{fig:full_system2}
        \end{subfigure}%
        \begin{subfigure}[b]{0.14\textwidth}
                \centering
                \includegraphics[scale=0.08]{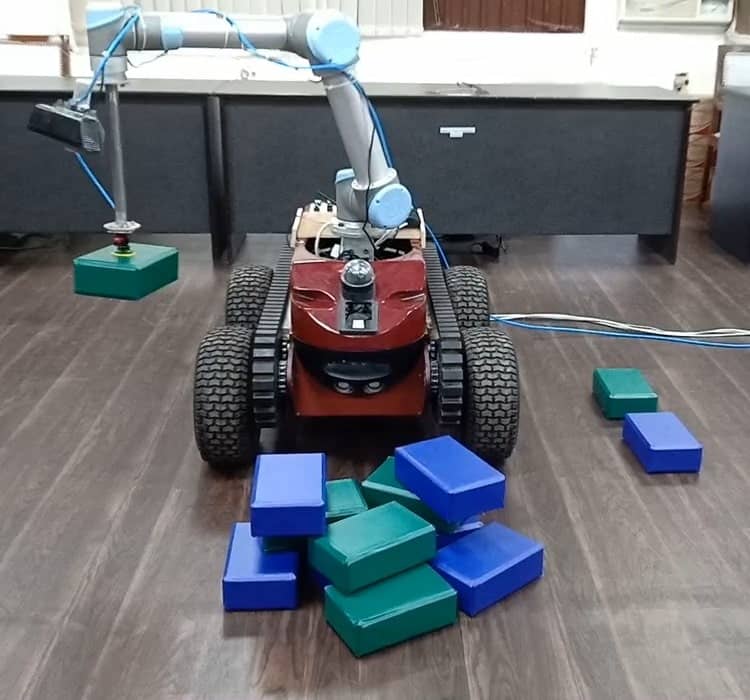}
                \caption{}
                \label{fig:full_system3}
        \end{subfigure}%
        \begin{subfigure}[b]{0.14\textwidth}
                \centering
                \includegraphics[scale=0.08]{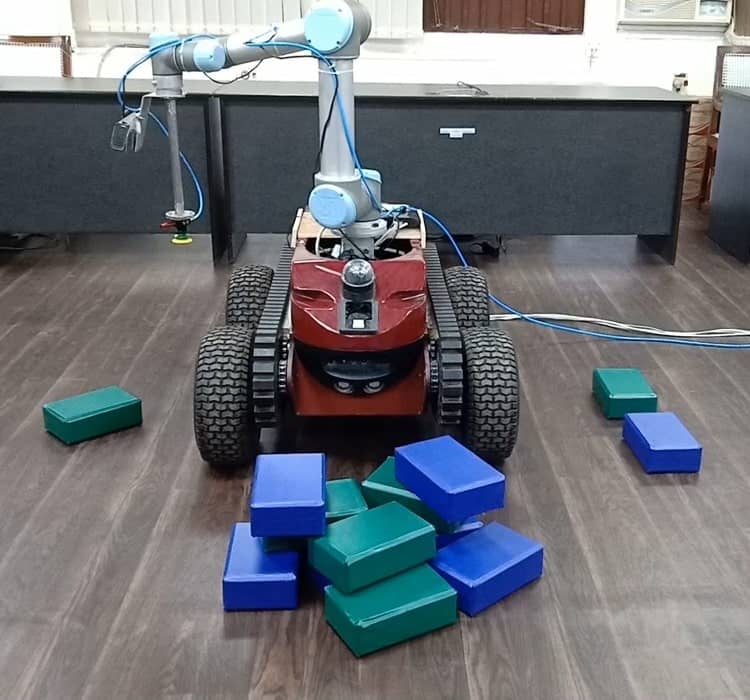}
                \caption{}
                \label{fig:full_system4}
        \end{subfigure}%
        \begin{subfigure}[b]{0.14\textwidth}
                \centering
                \includegraphics[scale=0.08]{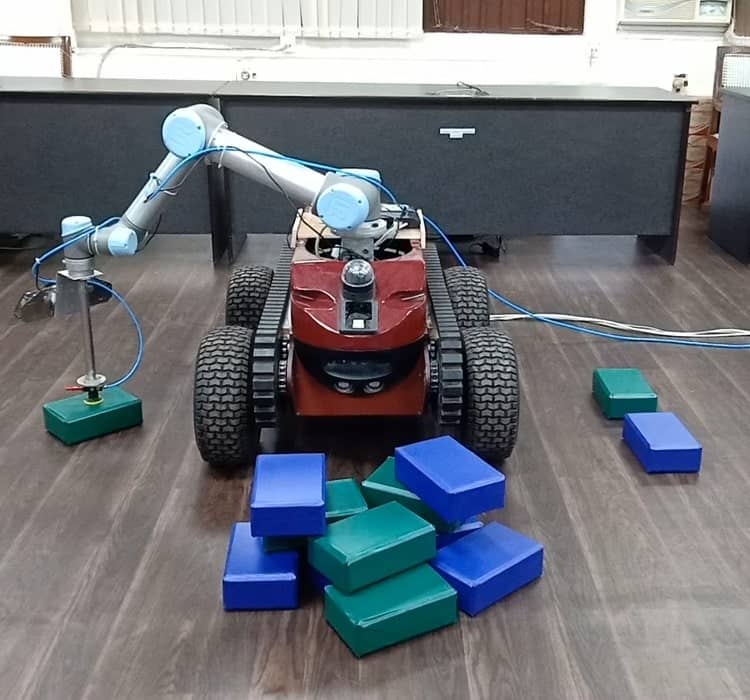}
                \caption{}
                \label{fig:full_system5}
        \end{subfigure}%
        \begin{subfigure}[b]{0.14\textwidth}
                \centering
                \includegraphics[scale=0.08]{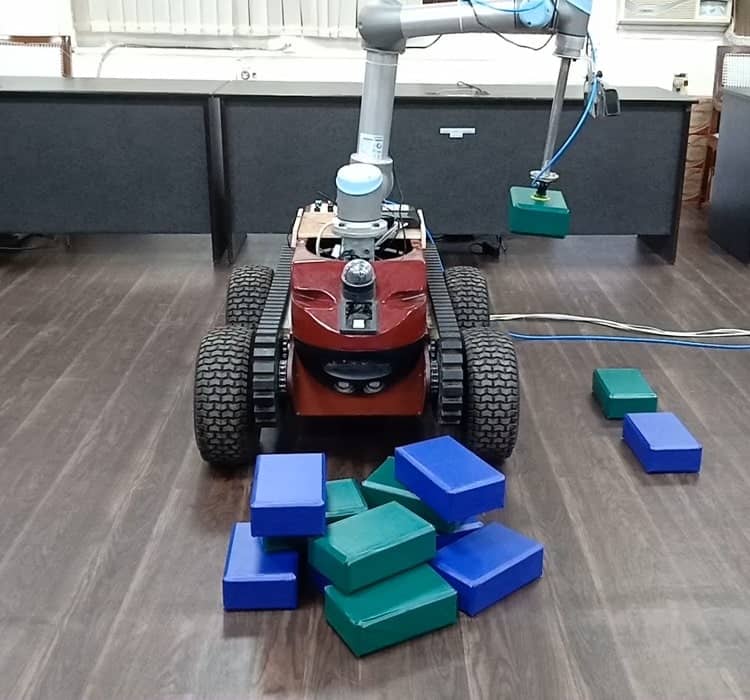}
                \caption{}
                \label{fig:full_system6}
        \end{subfigure}%
                \begin{subfigure}[b]{0.14\textwidth}
                \centering
                \includegraphics[scale=0.08]{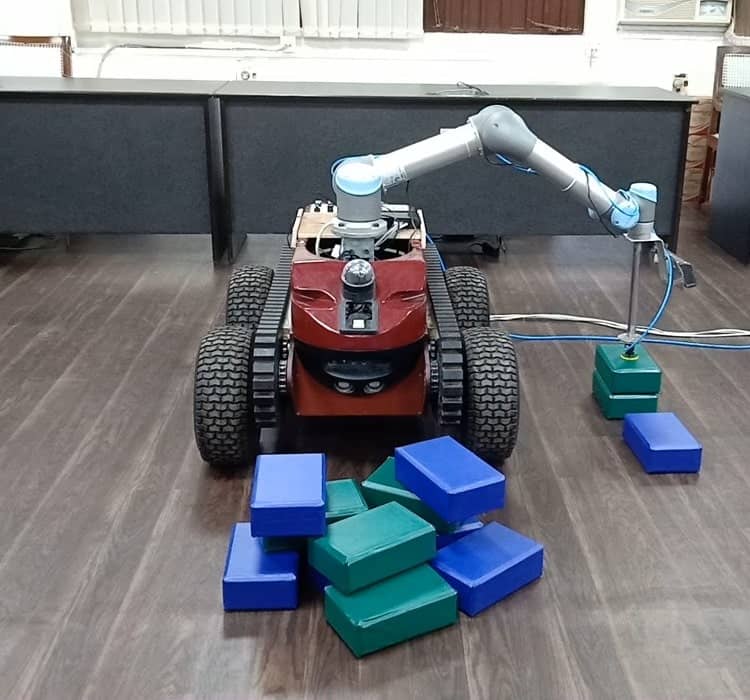}
                \caption{}
                \label{fig:full_system7}
        \end{subfigure}
        \caption{Sequence  of  actions  executed  in  order to carry out a single step of wall construction }
        \label{fig:full_system}
          \vspace{-4mm}

\end{figure*}

\subsection{Overall Algorithmic Flow}
For the experiment, we follow a simple pattern (or wall pattern) such that we place a blue brick on the previously placed blue brick and a green brick over the previously placed green brick and keep on placing the bricks up to the height of $6$ layers. The system needs to place the brick correctly for wall construction, hence requiring the brick pose with high accuracy. Since in a dense clutter, network prediction for brick can include some portion of other bricks, directly processing the box regions for pose can give a noisy or less reliable brick pose. For the safer side, we perform the grasp operation with a noisy brick pose and place the brick in a separate area and again estimate the pose of single isolated brick.

Fig. \ref{fig:full_system} refers to the sequence of steps executed to complete a single phase of the wall construction task. In the first stage, the Kinect sensor is positioned to have a clear view of the bricks clutter (Fig. \ref{fig:full_system1}), and the current image is fed to the rotating bounding box network. We compute the planar surface and its pose (centroid, major-axis, and minor-axis) corresponding to each rotated box. We select the topmost pose from all the calculated poses, i.e., the pose at the top of the clutter. The required motion commands are generated to reach the selected pose for brick grasping in clutter (Fig. \ref{fig:full_system2}).

The grasped brick is placed in a separate area (Fig. \ref{fig:full_system3}), and again the vision sensor is positioned in an appropriate position to have a clear view of a single isolated brick (Fig. \ref{fig:full_system4}). At this instant, the current frame is fed to the network, and the output of the network is a class label and the bounding box parameters. The brick's pose is estimated and required motion commands are generated to grasp the brick (Fig. \ref{fig:full_system5}).

In the final stage, the sensor is positioned in an appropriate position to have a view of the wall region (Fig. \ref{fig:full_system6}). The current frame is fed to the network. Using the estimated bounding box, point cloud data, and grasped brick label (previous step), the brick's final pose is estimated, where the brick is to be placed to form a wall (Fig. \ref{fig:full_system7}).

\subsection{Error Metrics}
The system's overall wall-building performance depends entirely upon the performance of the visual perception system, i.e., the accuracy of brick detection. Therefore, we report the performance of the detection system in terms of precision ($P$) and recall ($R$). We have defined $P,C$ as:
\begin{equation}
   P = \frac{NO}{TP},\ \ \  R = \frac{DB}{TB}
\end{equation}
\begin{description}
\item [where,]
\item [$NO$] Number of object pixels in the predicted box (rotated / upright)
\item [$TP$] Total number of object pixels in the predicted box (rotated / upright)
\item [$DB$] Total number of detected bricks
\item [$TB$] Total number of the bricks in the ground truth.
\end{description}

\subsection{Quantitative Analysis}
We compare rotating box network with YOLO-v3 and SSD-lite. For fair comparison, all models are trained by choosing hyper-parameters as $epoch=60, mini-batch=4, lr=0.0001$ except SSD-lite which has a learning rate of $0.001$. We use Adam optimizer to tune the parameters of CNN. We divide the quantitative analysis in following two cases:

\subsubsection{Upright Bounding Box Prediction}
In this case, we report Mean Average Precision (mAP) score, $P$ and $R$ of the proposed CNN model against SSD-lite and YOLO-v3 (Table-\ref{table_urb}). All the three models produces upright bounding boxes. 

\vspace{-7mm}
\begin{center}
\begin{table}[h]
\caption{}
\label{table_urb}
\centering
\begin{tabular}{| c | c | c | c | } 
\hline 
  & SSD-lite & YOLO-v3 & Proposed \\
\hline 
$P$ & $0.608$ & $0.580$ & $0.638$ \\
\hline 
$R$ & $0.98$ & $0.84$ & $0.84$ \\
\hline 
mAP & $0.834$ & $0.827$ & $0.811$ \\
\hline
\end{tabular}
\end{table}
\end{center}
\vspace{-7mm}

\subsubsection{Rotated Bounding Box Prediction}
In this case, SSD-lite and YOLO-v3 produce regular bounding boxes, while the proposed CNN model produces rotated boxes. Since the mAP score of rotated boxes can not be compared directly with that of upright bounding boxes. Hence only $P$ and $R$ is reported (Table-\ref{table_rb}). The precision $P$ for rotating boxes network is significantly higher as compared to other networks, because of the one additional degree of freedom (angle of boxes), the network predicts bounding boxes that can align more accurately as compared to constrained bounding boxes (straight boxes), thus there will be less overlap between different boxes and most of the region inside the bounding box represents the same object which results in high precision.

\vspace{-4mm}
\begin{center}
\begin{table}[h]
\caption{}
\label{table_rb}
\centering
\begin{tabular}{| c | c | c | c | } 
\hline 
  & SSD-lite & YOLO-v3 & Proposed \\
\hline 
$P$ & $0.608$ & $0.580$ & $0.778$ \\
\hline 
$R$ & $0.98$ & $0.84$ & $0.999$ \\
\hline
\end{tabular}
\end{table}
\end{center}
\vspace{-4mm}




\vspace{-4mm}
\subsection{Qualitative Analysis}
\begin{figure}

        \begin{subfigure}[b]{0.11\textwidth}
                \centering
                \includegraphics[scale=0.08]{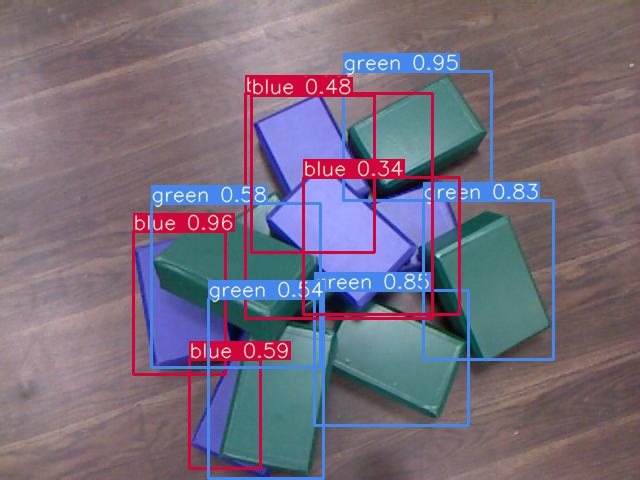}
                \caption{}
                \label{fig:all_output_1_1}
        \end{subfigure}%
        \begin{subfigure}[b]{0.11\textwidth}
                \centering
                \includegraphics[scale=0.08]{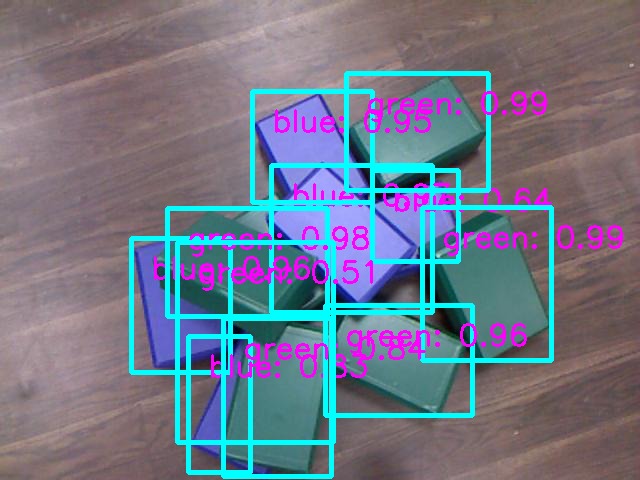}
                \caption{}
                \label{fig:all_output_1_2}
        \end{subfigure}%
        \begin{subfigure}[b]{0.11\textwidth}
                \centering
                \includegraphics[scale=0.08]{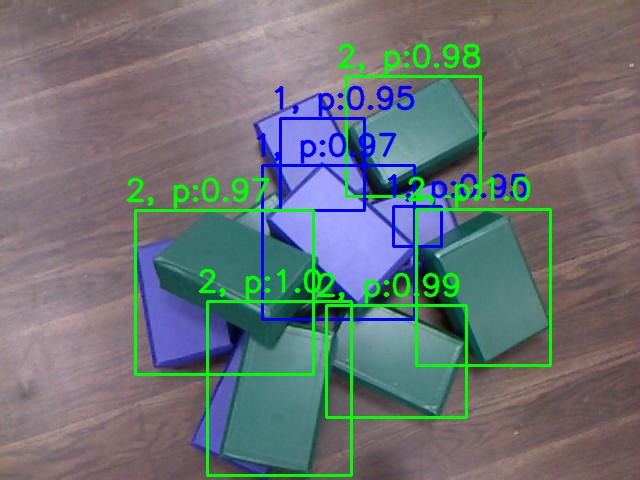}
                \caption{}
                \label{fig:all_output_1_3}
        \end{subfigure}%
        \begin{subfigure}[b]{0.11\textwidth}
                \centering
                \includegraphics[scale=0.08]{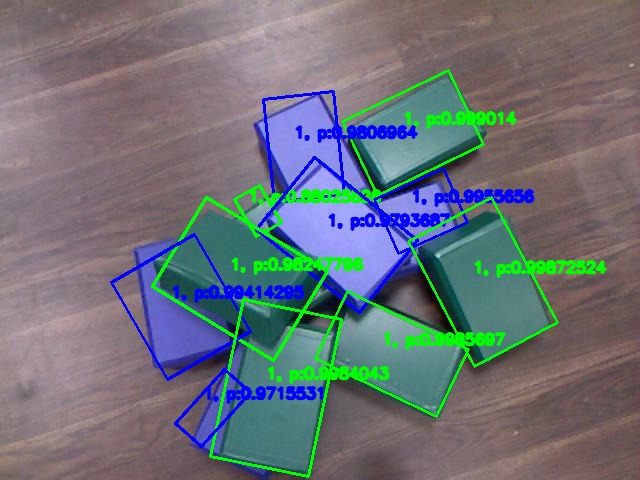}
                \caption{}
                \label{fig:all_output_1_4}
        \end{subfigure}
\medskip
        \begin{subfigure}[b]{0.11\textwidth}
                \centering
                \includegraphics[scale=0.08]{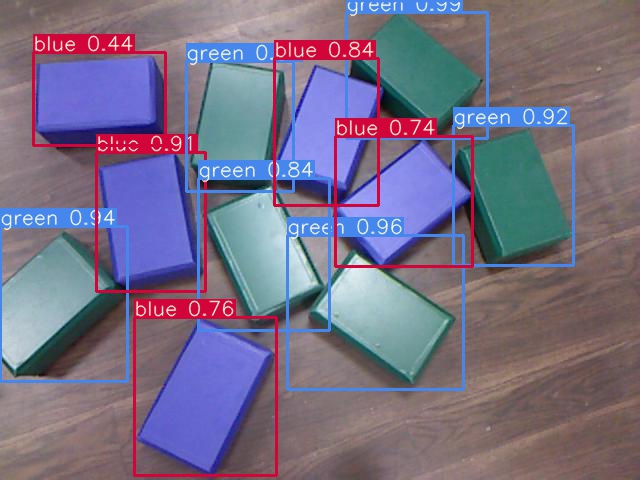}
                \caption{}
                \label{fig:all_output_2_1}
        \end{subfigure}%
        \begin{subfigure}[b]{0.11\textwidth}
                \centering
                \includegraphics[scale=0.08]{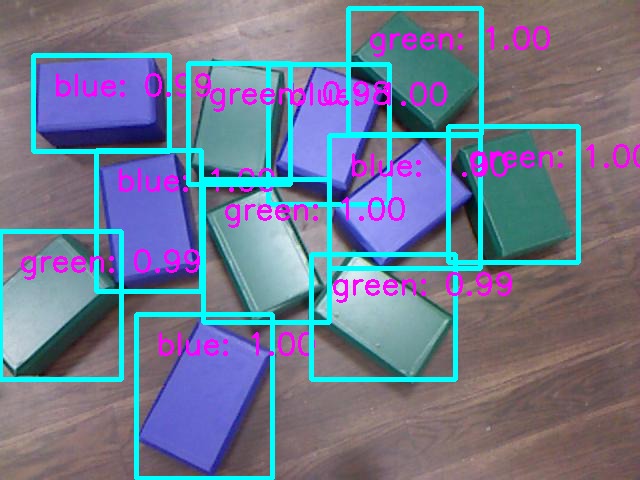}
                \caption{}
                \label{fig:all_output_2_2}
        \end{subfigure}%
        \begin{subfigure}[b]{0.11\textwidth}
                \centering
                \includegraphics[scale=0.08]{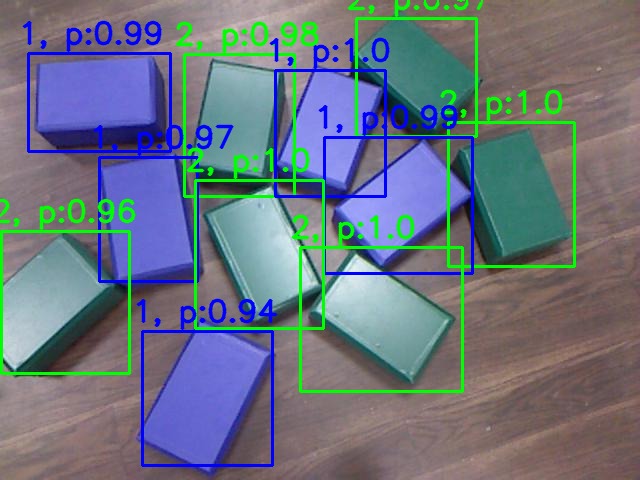}
                \caption{}
                \label{fig:all_output_2_3}
        \end{subfigure}%
        \begin{subfigure}[b]{0.11\textwidth}
                \centering
                \includegraphics[scale=0.08]{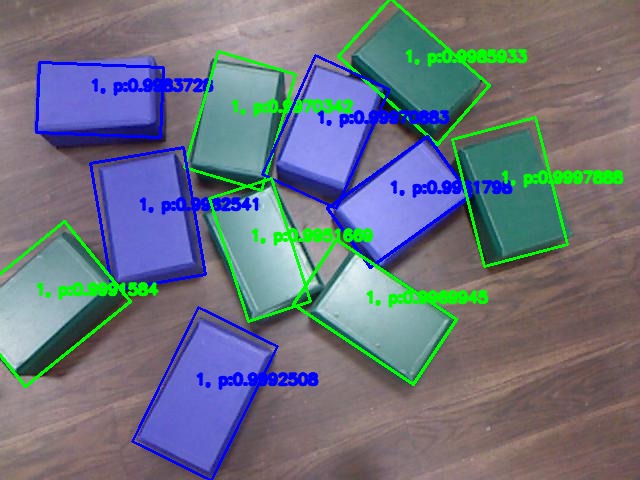}
                \caption{}
                \label{fig:all_output_2_4}
        \end{subfigure}
\medskip
        \begin{subfigure}[b]{0.11\textwidth}
                \centering
                \includegraphics[scale=0.08]{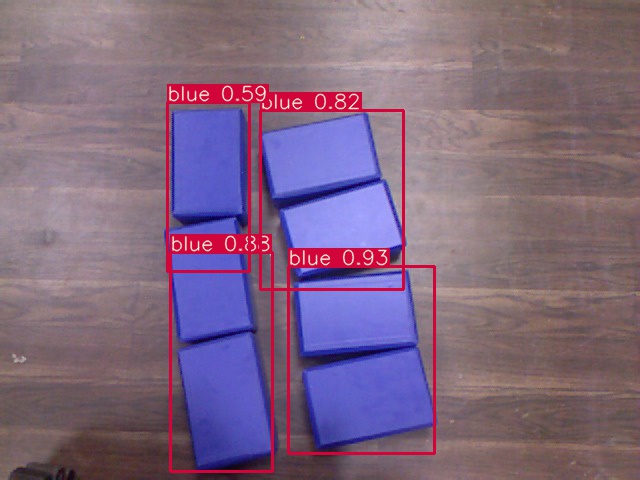}
                \caption{}
                \label{fig:all_output_3_1}
        \end{subfigure}%
        \begin{subfigure}[b]{0.11\textwidth}
                \centering
                \includegraphics[scale=0.08]{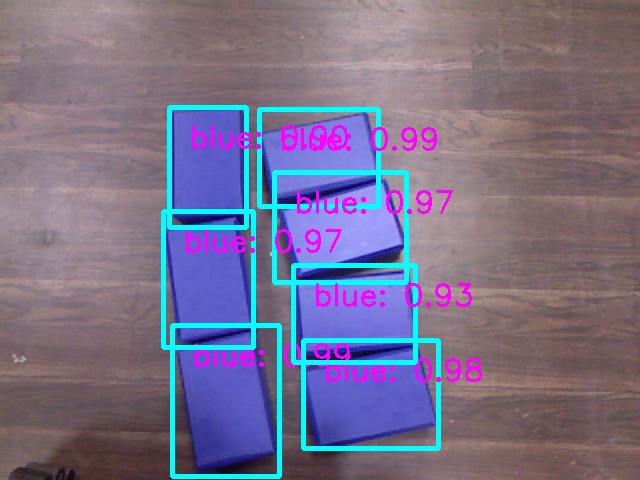}
                \caption{}
                \label{fig:all_output_3_2}
        \end{subfigure}%
        \begin{subfigure}[b]{0.11\textwidth}
                \centering
                \includegraphics[scale=0.08]{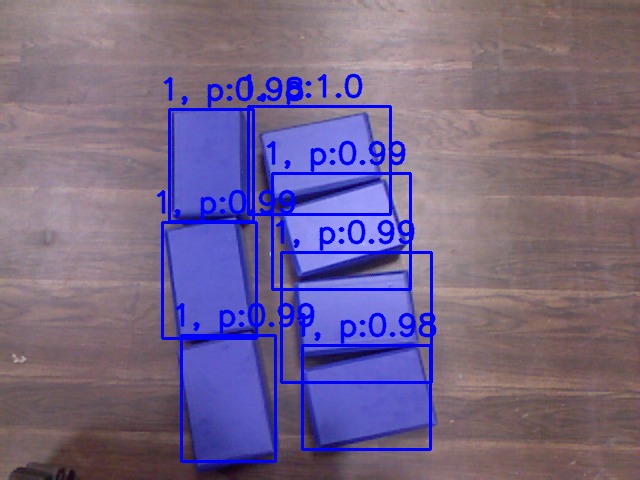}
                \caption{}
                \label{fig:all_output_3_3}
        \end{subfigure}%
        \begin{subfigure}[b]{0.11\textwidth}
                \centering
                \includegraphics[height = 0.66\textwidth, width =0.95\linewidth]{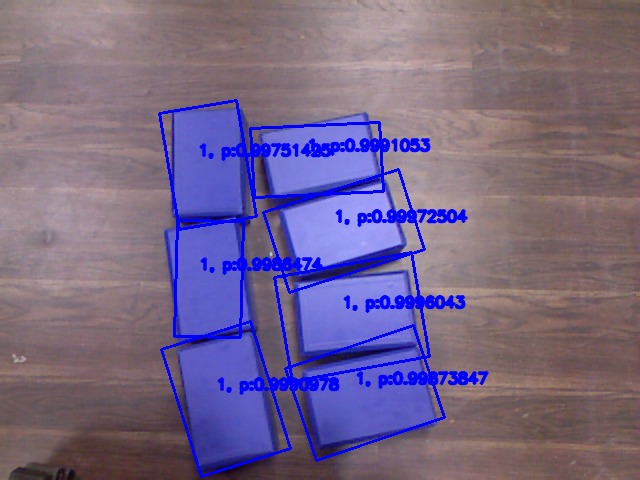}
                \caption{}
                \label{fig:all_output_3_4}
        \end{subfigure}
\medskip
        \begin{subfigure}[b]{0.11\textwidth}
                \centering
                \includegraphics[scale=0.08]{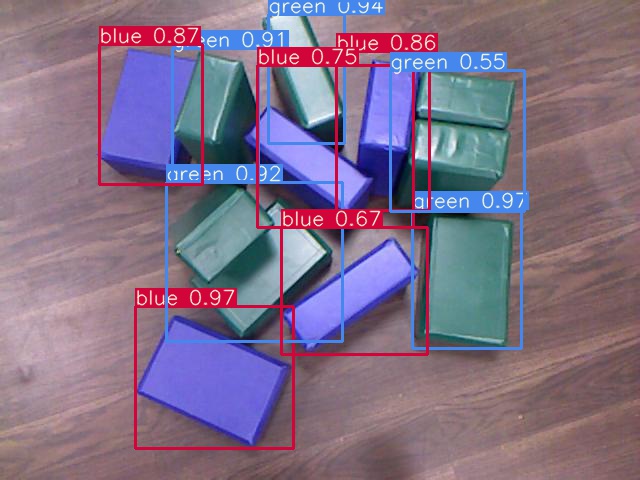}
                \caption{}
                \label{fig:all_output_4_1}
        \end{subfigure}%
        \begin{subfigure}[b]{0.11\textwidth}
                \centering
                \includegraphics[scale=0.08]{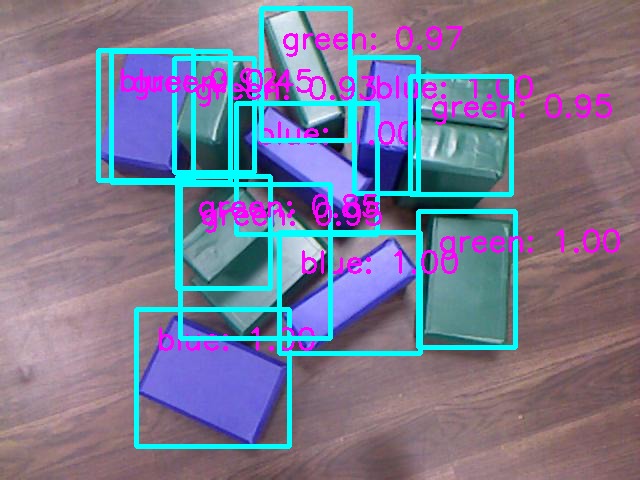}
                \caption{}
                \label{fig:all_output_4_2}
        \end{subfigure}%
        \begin{subfigure}[b]{0.11\textwidth}
                \centering
                \includegraphics[scale=0.08]{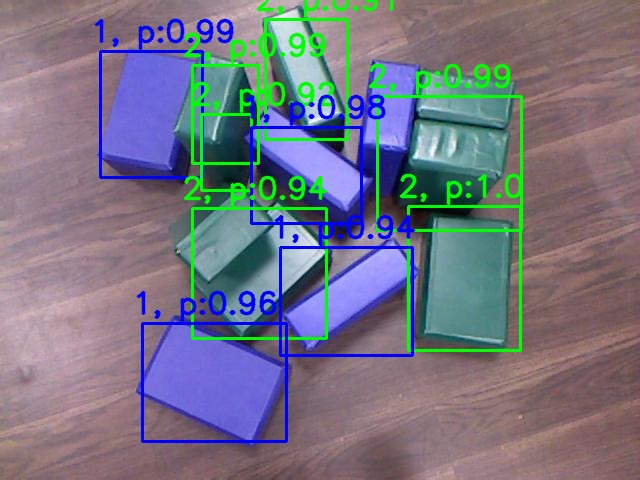}
                \caption{}
                \label{fig:all_output_4_3}
        \end{subfigure}%
        \begin{subfigure}[b]{0.11\textwidth}
                \centering
                \includegraphics[scale=0.08]{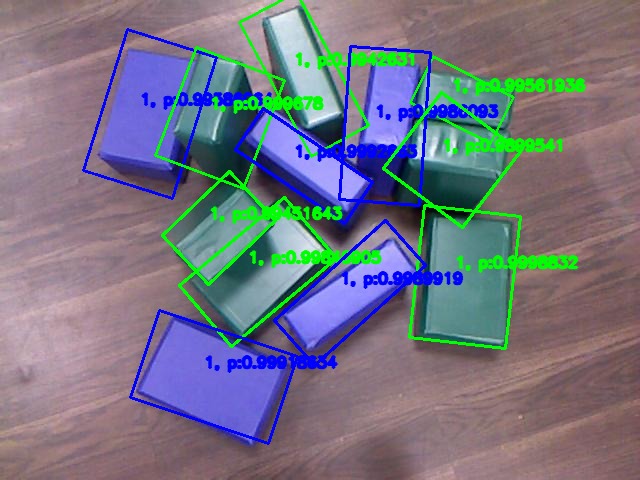}
                \caption{}
                \label{fig:all_output_4_4}
        \end{subfigure}
\medskip
        \begin{subfigure}[b]{0.11\textwidth}
                \centering
                \includegraphics[scale=0.08]{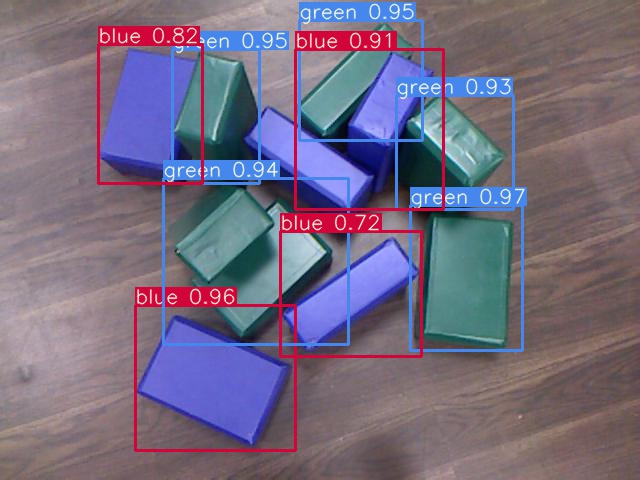}
                \caption{}
                \label{fig:all_output_5_1}
        \end{subfigure}%
        \begin{subfigure}[b]{0.11\textwidth}
                \centering
                \includegraphics[scale=0.08]{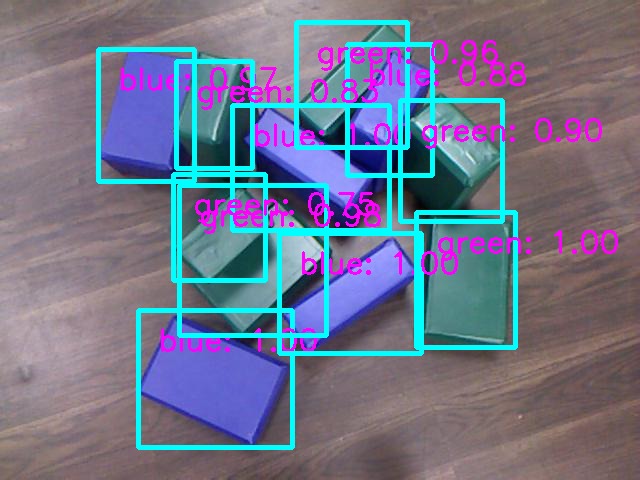}
                \caption{}
                \label{fig:all_output_5_2}
        \end{subfigure}%
        \begin{subfigure}[b]{0.11\textwidth}
                \centering
                \includegraphics[scale=0.08]{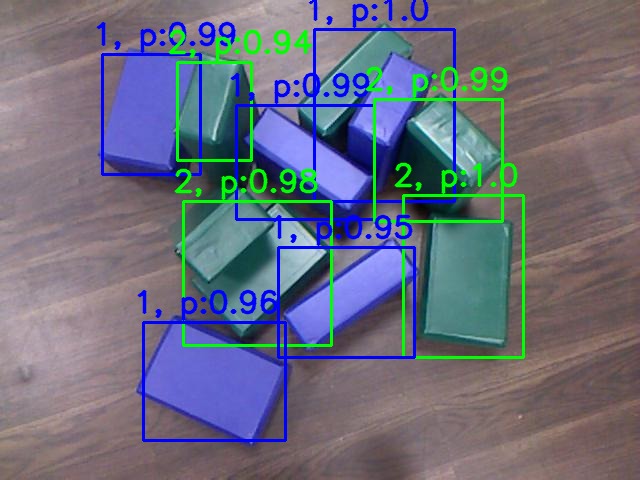}
                \caption{}
                \label{fig:all_output_5_3}
        \end{subfigure}%
                \begin{subfigure}[b]{0.11\textwidth}
                \centering
                \includegraphics[scale=0.08]{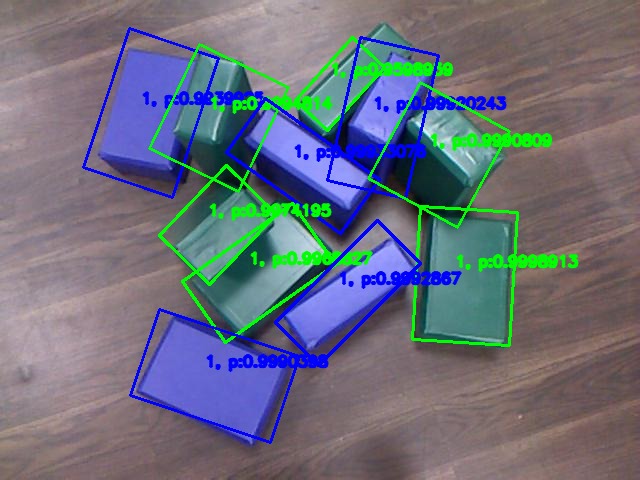}
                \caption{}
                \label{fig:all_output_5_4}
        \end{subfigure}
\caption{From Column 1 to 4: YOLO-v3 predictions, SSD-lite predictions, Proposed Bounding box predictions, Rotating box predictions.}
\label{fig:all_results}
\end{figure}

Predictions of all four networks under consideration are shown in Fig. \ref{fig:all_results}. It can be noticed from Fig. \ref{fig:all_output_4_1} and \ref{fig:all_output_4_3}, two green bricks present in the center of the frame are represented by a single bounding box, thus decreasing the recall value for Yolo-v3 and the proposed bounding box network. While SSD-lite (Fig. \ref{fig:all_output_4_2}) and the proposed rotating box network (Fig. \ref{fig:all_output_4_4}), both assign two different boxes for the two green bricks. Thus SSD-lite and our network have a higher recall. However, in SSD-lite, two predicted bounding boxes have a significant overlap area, thus having lower precision than the rotating box network.

\subsection{Task Evaluation} To evaluate the system's performance, we repeat the task of wall construction for $25$ rounds. For each round, the robotic system has to place the bricks up to the height of $6$ layers, and the wall pattern will remain the same, i.e., the blue brick on previously placed blue brick and green brick on previously placed green brick. For the experiments, the first layer of bricks is placed manually, and the system has to place the rest of the bricks, i.e., $2-6$ layers, according to the pattern. We count the number of bricks (or layers) for each round the robotic system has successfully placed on the wall. In the experiments, we define the brick placement as successful if the distance between the centroid of the currently placed brick and the centroid of the manually placed brick, when projected on the ground plane is $< 0.1m$. Further, the Euler angle difference between the calculated pose of the currently placed brick and the manually placed brick should be less than \ang{15} for each axis. From our experiments, we observed that if none of the above criteria are satisfied, the wall becomes asymmetrical and collapses. The Table-\ref{table_task_eval} shows the performance of the system for $25$ rounds. The video link for the experiment is \url{https://www.youtube.com/watch?v=FvsCv-Pt58c}.
\vspace{-4mm}
\begin{center}
\begin{table}[h]
\caption{Task Evaluation}
\label{table_task_eval}
\centering
\begin{tabular}{ | C{1.2cm} | C{0.8cm}| C{0.8cm} | C{0.8cm} | C{0.8cm} | C{0.8cm} | } 
\hline 
  & layer-2  & layer-3 & layer-4 & layer-5 & layer-6 \\
 \hline 
Successful rounds (max 25) & 25  & 25 & 22 & 19 & 17 \\
 \hline 
\end{tabular}
  \vspace{-4mm}

\end{table}
\end{center}
\vspace{-4mm}

From Table-\ref{table_task_eval}, we observed that the robotic system has successfully placed layer-$2$ and layer-$3$ bricks in all $25$ rounds. However, accuracy decreases with the upper layers. This is because the new position of the brick on the wall is estimated by calculating the previously placed brick pose. Thus, if there is a slight error in brick placement in the previous step, this error is transferred to the current step. Thus, with higher layers, the error accumulates, resulting in lower accuracy.

\section{Conclusion}
\label{sec_con}
An end-to-end visual perception framework is proposed. The framework consists of a CNN for predicting a rotated bounding box. The performance of the CNN detector has been demonstrated in various scenarios, which mainly include isolated and dense clutter of bricks. The proposed CNN module localizes the bricks in a clutter while simultaneously handling multiple instances of the bricks. The detection is free of the anchor-box technique, which improves the timing performance of the detection module. In order to compare our method quantitatively with state-of-the-art models, we reported Precision ($P$), Recall ($R$), and mAP scores for various test cases. We have compared the effectiveness of rotating bounding box predictions against upright bounding box detection (YOLO-v3, SSD-Lite). The proposed scheme outperforms the upright bounding box detection. It implies that rotating bounding boxes can align more accurately with the object's convex-hull and thereby reduce the overlap with neighboring bounding boxes(if any). The framework has also been successfully deployed on a  robotic system to construct a wall from bricks in fully autonomous operation.

\bibliography{citations} 
\bibliographystyle{ieeetr}
\end{document}